\documentclass[AMA,STIX1COL]{WileyNJD-v2}

\articletype{Special Issue Paper}%

\received{<day> <Month>, <year>}
\revised{<day> <Month>, <year>}
\accepted{<day> <Month>, <year>}

\begin{document}

\title{A multi-domain VNE algorithm based on multi-objective optimization for IoD architecture in Industry 4.0}

\author[1]{Peiying Zhang}

\author[1]{Chao Wang}

\author[2]{Zeyu Qin}

\author[3]{Haotong Cao}

\authormark{ZHANG \textsc{et al}}

\address[1]{\orgname{College of Computer Science and Technology}, \orgdiv{China University of Petroleum (East China)}, \orgaddress{\state{Qingdao}, \country{China.}}}


\address[2]{\orgname{State Key Laboratory of Networking and Switching Technology}, \orgdiv{Beijing University of Posts and Telecommunications},  \orgaddress{\state{Beijing}, \country{China.}}}

\address[3]{\orgname{College of Telecommunications and Information Engineering}, \orgdiv{Nanjing University of Posts and Telecommunications},  \orgaddress{\state{Nanjing}, \country{China.}}}

\corres{Haotong Cao, Nanjing University of Posts and Telecommunications, Nanjing, China. \email{1015010309@njupt.edu.cn} \\
Peiying Zhang, China University of Petroleum (East China), Qingdao, China. \email{25640521@qq.com}
~\\
~\\
\textbf{Funding Information}
\\This work is supported by "the Fundamental Research Funds for the Central Universities" of China University of Petroleum (East China) (Grant No. 18CX02139A), the Shandong Provincial Natural Science Foundation, China (Grant No. ZR2014FQ018), and the Demonstration and Verification Platform of Network Resource Management and Control Technology (Grant No. 05N19070040).}


\abstract[Abstract]{Unmanned aerial vehicle (UAV) has a broad application prospect in the future, especially in the Industry 4.0. The development of Internet of Drones (IoD) makes UAV operation more autonomous. Network virtualization technology is a promising technology to support IoD, so the allocation of virtual resources becomes a crucial issue in IoD. How to rationally allocate potential material resources has become an urgent problem to be solved. The main work of this paper is presented as follows: (1) In order to improve the optimization performance and reduce the computation time, we propose a multi-domain virtual network embedding algorithm (MP-VNE) adopting the centralized hierarchical multi-domain architecture. The proposed algorithm can avoid the local optimum through incorporating the genetic variation factor into the traditional particle swarm optimization process. (2) In order to simplify the multi-objective optimization problem, we transform the multi-objective problem into a single-objective problem through weighted summation method. The results prove that the proposed algorithm can rapidly converge to the optimal solution. (3) In order to reduce the mapping cost, we propose an algorithm for selecting candidate nodes based on the estimated mapping cost. Each physical domain calculates the estimated mapping cost of all nodes according to the formula of the estimated mapping cost, and chooses the node with the lowest estimated mapping cost as the candidate node. The simulation results show that the proposed MP-VNE algorithm has better performance than MC-VNM, LID-VNE and other algorithms in terms of delay, cost and comprehensive indicators.}



\maketitle

\footnotetext{\textbf{Abbreviations:} ANA, anti-nuclear antibodies; APC, antigen-presenting cells; IRF, interferon regulatory factor}

\section{Introduction}\label{sec1}

In industry 4.0\cite{Liu2019How}, countries all over the world pursue the development of advanced science and technology, among which UAV has a broad development prospect. Drones are widely used in military, agricultural and now popular logistics robots. But cost, technology and other factors limit the development of UAV technology. The development of IoD enables UAV to operate autonomously as much as possible. IoD focuses on computing, communications, security and privacy, energy and sustainability\cite{Wu2019Research}. The limited computing resources that UAVs can provide must be effectively utilized in processing, computing, perception and analysis to remain lightweight. These high performance requirements are very strict for grassroots networks. UAV technology requires the network to provide safe, reliable, high-speed and convenient services. Traditional network system can not effectively handle IoD dynamic requirements. The latest network models, such as software defined network (SDN) and network function virtualization (NFV)\cite{Cao2019An}, offer new potential for technical advances in IoD communication architecture. NFV technology can be understood as a new network architecture, its quality of service is more reliable, faster, and can cope with the high performance requirements of UAV network. Radio network resource management faces severe challenges, including storage, spectrum, computing resource allocation, and joint allocation of multiple resources \cite{jcx1,jcx2}. With the rapid development of communication networks, the integrated space-ground network has also become a key research object \cite{jcx3}.

The development of the Internet has provided convenience for people\cite{Cao2019Dynamic}. However, with the introduction of new protocols and technologies, the Internet has become bloated and rigid. The Internet architecture only provides "best efforts" delivery\cite{http2015Internet, Kurose2013Computer, Porter2013Internet} which cannot meet users' demand for service diversity\cite{Wu2008Fundamental, Moreno2013Key, Anderson2014A, Bechtold2014Accountability, Fisher2014A}. In response to this problem, T. Anderson et.al.\cite{T2005Overcoming} proposed network virtualization technology, which allows various protocols to coexist in a shared substrate infrastructure. Besides, it can reasonably allocate network resources according to dynamically changing user requirements\cite{Chowdhury2009Network, Khan2012Network, Carapinha2009Network}. Network virtualization technology is considered to be one of the most promising technologies in the future Internet\cite{Ines2010Virtual}.

Virtual local area networks (VLAN) and virtual private networks (VPN) are examples of network virtualization. The primary goal of network virtualization is to build a robust, trusted, and manageable virtual environment. Allocate appropriate virtual resources for various virtual network requests, realize resource sharing and improve utilization of infrastructure resources. VPN is a technology which makes full use of tunnel, authentication, encryption and so on. It can enable users to access resources safely and remotely. Similar to the VPN problem, virtual network mapping allocates resources to the logical topology with resource constraints in the Shared underlying network. However, a typical VPN request only contains link resource constraints, while in the virtual network mapping problem, each virtual network request contains not only link resource constraints, but also node constraints such as computing power and location requirements\cite{Han2011Research, Xu2004Comparison,Cheng2011Research}.

At present, single-domain virtual network mapping is increasingly unable to meet people's business needs\cite{Cao2018Novel}, because in practical applications, we often need to comprehensively consider multiple indicators of the mapping results. IoD is one of the core technologies supporting the development of UAV. Traditional network systems may not be very effective in dealing with the dynamic requirements of IoD. The latest network models, such as software-defined virtualization of network and network functions, provide new potential for the technological advancement of IoD communication architecture. Therefore, the study of network virtualization is of great significance to the development of IoD.

The main innovation points of this paper are summarized as follows:

\begin{enumerate}
\item In order to improve the optimization performance and aiming at the problem of computing speed is not high. We propose a virtual network mapping algorithm based on particle swarm optimization MP-VNE. The algorithm adopts the centralized hierarchical multi-domain mapping virtual network architecture. MP-VNE algorithm can collect each domain information and comprehensive treatment. The genetic variation factor is introduced into the traditional particle swarm optimization (PSO) algorithm, which can effectively avoid the situation that the particle falls into the local optimum but cannot reach the global optimum.
\item In order to make the algorithm converge to the optimal solution as soon as possible, the multi-objective problem is transformed into a single objective problem by using the weighted value.
\item In order to save mapping cost and select physical nodes quickly, we propose a candidate node selection algorithm based on estimated mapping cost. Each physical domain calculates the estimated mapping cost of all nodes according to the formula proposed, and selects the node with the lowest estimated mapping cost as the candidate node.
\end{enumerate}

The rest of this paper is arranged as follows. Chapter 2 introduces the research status of multi-domain virtual network mapping. Chapter 3 describes the problem of virtual network mapping. Chapter 4 introduces the design of multi-domain virtual network mapping algorithm. Chapter 5 elaborates on the implementation of multi-domain virtual network mapping algorithm in detail. Chapter 6 calculates it. The simulation results of the method are described and analyzed. Chapter 7 summarizes the paper.

\section{Related Works}
Because of the distributed characteristics of the IoD architecture, it has many advantages in scalability and autonomy. This advantage makes the design of new architecture, new standard and new technology of UAV more independent, but also brings a variety of mobility and control problems. The traditional network system may not be very effective in dealing with the dynamic demand timing of IoD. The latest network models, such as the virtualization of network functions, provide new possibilities for the technological progress of the communication architecture of IoD.

In the current centralized multi domain virtual network mapping architecture, the centralized server as the central node plays the role of manager. It receives virtual network mapping requests, collects information from various infrastructure providers, integrates the collected information and determines the virtual network mapping program\cite{Cao2018A, Zhang2014Virtual, Walter2001A, Jiang2002A}.

\subsection{Centralized Multi-domain Virtual Network Mapping}
Currently, many algorithms adopt centralized multi-domain virtual network mapping architecture to improve optimization objectives. For example, Peng Limin et al\cite{Peng2015Cross-domain}, proposed a multi-domain virtual network mapping algorithm based on Kruskal minimum spanning tree. This scheme chooses the minimum weight physical path in turn from the set of available mapping physical paths, then maps the corresponding virtual link to the physical path and coordinates the mapping operation of virtual nodes. Geng Ruiwen et al\cite{Geng2016Multi-domain}, proposed a multi-domain virtual network mapping algorithm based on mapping overhead estimation. In this scheme, firstly, candidate nodes in the domain are selected according to candidate node selection algorithm based on node and link mapping overhead estimation. Then the local controller uploads candidate nodes to the global controller. Finally, the global controller uses particle swarm optimization algorithm to select the virtual network. The network request is pre-mapped and the pre-mapped result is sent to the corresponding local controller for mapping. Xiao Ailing\cite{Xiao2014Knowledge} and others proposed a multi-domain virtual network mapping algorithm based on knowledge description and genetic algorithm. They designed a resource matching algorithm based on Web Ontology Language (OWL) knowledge description and Semantic Web Rule Language (SWRL) query tool and proposed a virtual network (VN) partition algorithm based on genetic algorithm.

\subsection{Distributed Multi-domain Virtual Network Mapping}
The architecture of IoD has distributed characteristics. There is no centralized central control node in the distributed multi-domain virtual network architecture to maintain the global state information of the physical network. The final virtual network mapping scheme is determined by multiple information exchanges between service providers and infrastructure providers. For example, Samuel F \cite{SamuelF2013PolyViNE} and others proposed a mapping strategy based on bidding mechanism. This strategy sends the virtual network mapping request to an InP. When InP receives the virtual network mapping request, it selects the subgraph of the virtual network mapping request to be embedded and passes the rest of the virtual network mapping request to other InPs. Dietrich $et$ $al.$ \cite{Dietrich2013Multi-domain} proposed a mapping strategy based on matrix decomposition. In this scheme, a virtual network mapping request is modeled as a traffic matrix. The physical domain of virtual node mapping is determined first, then the traffic matrix is decomposed into several small matrices and sent to the corresponding physical domain. Each physical domain independently carries out nodes and chains according to the received small matrices. Mapping the path and sending the local quotation to SP. The authors of \cite{Zaheer2010Multi-provider} proposed a V-Mart mechanism based on Vickrey auction model. V-Mart adopts a two-stage Vickey auction model, which can be flexibly applied to various InP pricing models and play a better role in multi-commodity markets.

\subsection{Summary of This Chapter}
The centralized multi-domain virtual network mapping architecture can obtain the state information of each physical domain. However, the efficiency of virtual network mapping will decrease in larger scale physical network. The distributed multi-domain virtual network mapping architecture has good scalability and high computational efficiency. However, there is no central control node that understands the global resource situation, which makes it difficult to find a suitable mapping scheme.

We propose a centralized multi-layer virtual network mapping architecture. As with the basic centralized architecture, each physical domain has a local controller that implements resource statistics and virtual network mapping. The global controller is responsible for receiving virtual network mapping requests and information uploaded by the local controller, then making the final mapping decision. The difference is that when the global controller makes the mapping scheme decision, the particle swarm optimization algorithm with mutation factor is adopted. Avoid falling into the condition of local optimum and not reaching global optimum.

\section{Multi-domain Virtual Network Mapping Problem Description}

\subsection{Multi-domain Virtual Network Mapping Problem}
Virtual network mapping technology mainly solves the problem of mapping a virtual network request to the underlying physical network, namely the allocation of resources of physical nodes and physical links\cite{M2012Network, Lv2011Research, Sunl2012Research}. It is divided into two steps: Firstly, virtual node mapping, in which each virtual node is mapped to a single physical node. Secondly, virtual link mapping, in which each virtual link is mapped to one or more virtual links\cite{Houidi2008A}.

There are two main roles in network virtualization: Infrastructure Provider (InP) and Service Provider (SP)\cite{Werle2011Building}. In the problem of virtual network mapping, SP obtains the virtual network mapping request from the user, then according to the information provided by InP, generates a reasonable virtual network mapping scheme to map the virtual network request to the underlying network.

The basic virtual network mapping only considers mapping in a single-domain environment, but the actual deployed virtual network may go through multiple administrative domains or data centers across the underlying infrastructure. The mapping problem of multi-domain virtual network is closer to reality than that of basic virtual network, but it is also more complicated. Multi-domain virtual network mapping requires that virtual networks can be embedded, deployed and operated across multiple InPs. This is a major challenge compared with the single domain virtual network mapping problem, mainly because InP's policy restricts the disclosure of resource information and hinders the inter-operability of resources with other parties. However, the single-domain virtual network mapping algorithm usually needs to fully understand the underlying network resource distribution and network topology\cite{Zhu2016Algorithms, M2008Rethinking}. Therefore, the single-domain virtual network mapping algorithm is not suitable for the scenario of multi-domain virtual network mapping.

\subsection{Problem Modeling}

The mapping process based on the centralized hierarchical multi-domain virtual network mapping architecture is completed by the global controller and the local controller. The role of the global controller is to receive virtual network requests and decompose virtual network map requests into subgraphs. The global controller can also send subgraphs to each local controller and make the final virtual network mapping according to the information uploaded by each local controller. The local controller is responsible for uploading the virtual network information in this controller to the global controller. According to the subgraph issued by the global controller, candidate nodes of each virtual node are selected and uploaded to the global controller for the mapping of virtual network. The local controller can also map nodes and links according to the final scheme issued by the global controller.

The entire physical domain is modeled as a weighted undirected graph, $G^s=({G_i^s},L^s)$, a collection containing a single physical domain ${G_i^s}$ and inter-domain links $L^s$.

Single physical domain $G_i^s=(N_i^s,L_i^s)$ is also a powerful undirected graph, $N_i^s$ is a collection of physical nodes, $L_i^s$ is a collection of links in a domain.

$N_i^s=(C_{Ni}^s,P_{Ni}^s,D_{Ni}^s)$, each physical node $n_i^s$ belongs to $N_i^s$ has three attribute values: $C_{Ni}^s$ is the CPU processing capacity of physical nodes, that is, the resource quantity of nodes. The more resources provided by CPU for virtual node and virtual link mapping in unit time, the stronger its processing power. $P_{Ni}^s$ is the unit price of resources of the physical node. $D_{Ni}^s$ is the time required by physical nodes to process unit resources, including node processing delay, queuing delay, transmission delay, etc.

$L_i^s=(B_{Ni}^s,P_{Ni}^s,D_{Ni}^s)$, each intra-domain link also has three attribute values: bandwidth capacity($B_{Li}^s$), price($P_{Li}^s$), time delay($D_{Li}^s$). $B_{Li}^s$ is the carrying capacity of the physical link. $P_{Li}^s$ is the resource unit price of physical link. $D_{Li}^s$ is the time required by physical link to transmit information, including propagation delay. The same is true for inter-domain links.

\begin{figure}[!htbp]
\centerline{\includegraphics[width=342pt,height=18pc]{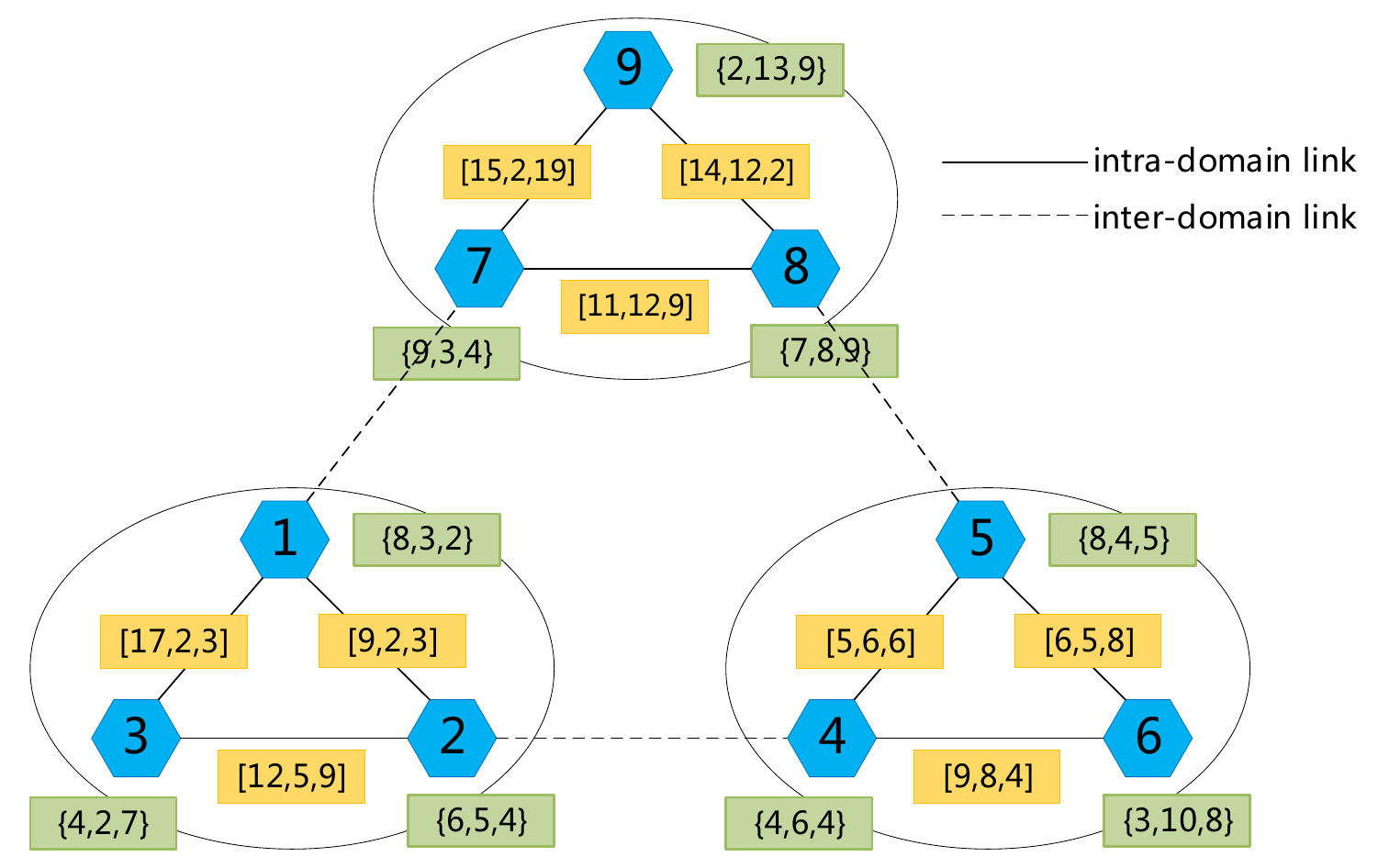}}
\caption{The substrate network example.}
\label{fig_physical network extension}
\end{figure}

In FIGURE \ref{fig_physical network extension}, each circle represents a physical node. The number in the circle is the number of the physical node. The number in parentheses on the circle is the basic information of the physical node, which are respectively CPU processing power, unit price of node resources and node delay. The ellipse represents a physical domain. The physical nodes in a single physical domain are connected by solid lines, representing the links in the domain. The nodes between different physical domains are connected by dotted lines, representing the links between domains. The Numbers in brackets beside the links represent the link carrying capacity, link resource unit price and link propagation delay respectively. Three ellipses represent three physical domains, which can well embody the concept of cross-domain. Each physical domain has multiple switches or physical nodes connected together to form a physical network. The physical networks are connected by inter-domain links. The nodes connected with other fields are called boundary nodes. In the FIGURE \ref{fig_physical network extension}, nodes 1, 2, 4, 5, 7 and 8 are boundary nodes.

A virtual network is defined as a entitled undirected graph $G^v=(N^v,L^v)$. $N^v$ is a collection of virtual nodes, $L^v$ is a collection of virtual links.

$N^v=(C^v,D^v)$, each virtual node contains two properties. $C^v$ represents the CPU demand of the virtual node. $D^v$ represents a candidate domain for a virtual node that can only be mapped to a specified domain. $L^v=(B_l^v)$, each virtual link contains only one property. $B_l^v$ represents bandwidth requirements.


The goal of this algorithm is to take into account the mapping cost and network delay, so that the two comprehensive indicators are optimal. This comprehensive indicator is called Cost. The objective function is:
\begin{equation}
\begin{aligned}
Cost=\sum_{Num(n^v)}CPU(n^v) \times Cost(n_v^s)+\sum_{Num(l^v)}bw(l^v) \times Cost(l_v^s).
\end{aligned}
\end{equation}

Among them, CPU($n^v$) is the resource demand of a node in the virtual network request; bw($l^v$) is the bandwidth requirement of a link in the virtual network request; $n_v^s$ represents the final mapping of the virtual node; $l_v^s$  represents the final mapping of virtual links; Cost($n_v^s$) is the unit price of physical nodes; Cost($l_v^s$) represents the physical path link aggregation of unit price. If the physical path is an intra domain path, the link aggregation unit price is the sum of the link unit prices between the two physical nodes in the physical domain. If this path is a cross domain path, then the unit price of resource aggregation is the unit price synthesis of the link from the start and end of the two physical domains to the boundary node and the cross domain link between the two physical domains.

\section{Design of Multi-Domain Virtual Network Mapping Algorithm}

\subsection{Algorithm Core Idea}
This paper proposes a centralized multi-layer virtual network mapping architecture based on single objective method and discusses the possibility of centralized features promoting the technological progress of the IoD architecture. In this scheme, each physical domain has a local controller to implement the statistics of resources in the physical domain and the mapping of virtual nodes and links. In addition, there is a global controller for receiving virtual network mapping requests and information uploaded by local controllers. The local controller selects candidate nodes according to the candidate node selection algorithm and uploads the generated information to the global controller to determine the final mapping scheme. In the decision-making process, the particle swarm optimization algorithm with mutation factor is adopted.

\subsection{Algorithm Description}

The multi-domain virtual network mapping algorithm proposed can be roughly divided into two parts. The division and overall optimization of virtual network mapping requests by the global controller and the specific embedding of virtual network requests by the local controller. Its flow chart is shown in FIGURE 2. In the figure, the global controller flowchart is on the left and the local controller flowchart is on the right.

\begin{figure}[!htp]
\centerline{\includegraphics[width=350pt,height=25pc]{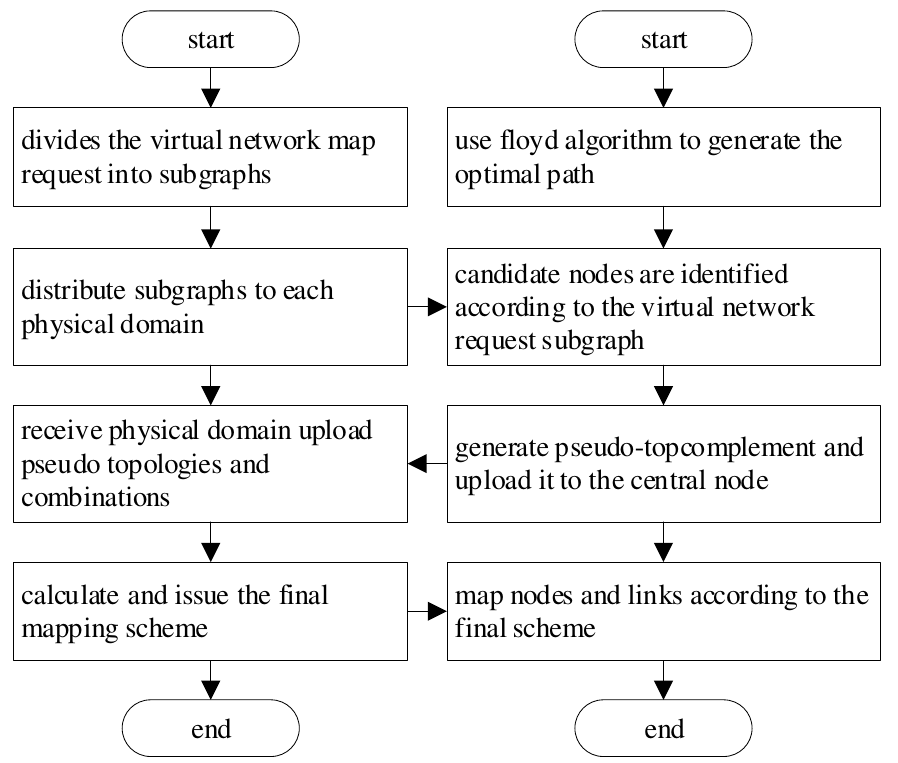}}
\caption{The MP-VNE algorithm flow.}
\label{fig_MP-VNE algorithm flow}
\end{figure}

\begin{enumerate}
\item After receiving the virtual network request from SP, the global controller divides it into several subgraphs. According to the local network topology information, the local controller uses Floyd algorithm to find the optimal route.
\item The global controller sends the subgraph of virtual network request unit to the local controller. The local controller calculates the candidate node set according to the information in the domain.
\item The local controller integrates candidate nodes, boundary nodes and other information into a pseudo-topology and uploads it to the global controller. The global controller combines the received pseudo-topologies to produce a large cross-domain pseudo-topology.
\item The global controller is iteratively optimized according to particle swarm optimization algorithm on this large cross-domain pseudo-topology. Its objective function is to estimate the mapping cost, calculate the optimal solution under the current condition and distribute the mapping scheme to the local controller. According to this mapping scheme, the local controller maps in the domain and calculates the cost of upload mapping.
\item If all local controllers are successfully mapped, the virtual network mapping request is successfully mapped. If the local controller mapping fails (for reasons such as insufficient resources), the mapping fails.
\end{enumerate}

\subsection{Algorithm Function Module}
\begin{figure}[!htp]
\centerline{\includegraphics[width=430pt,height=13pc]{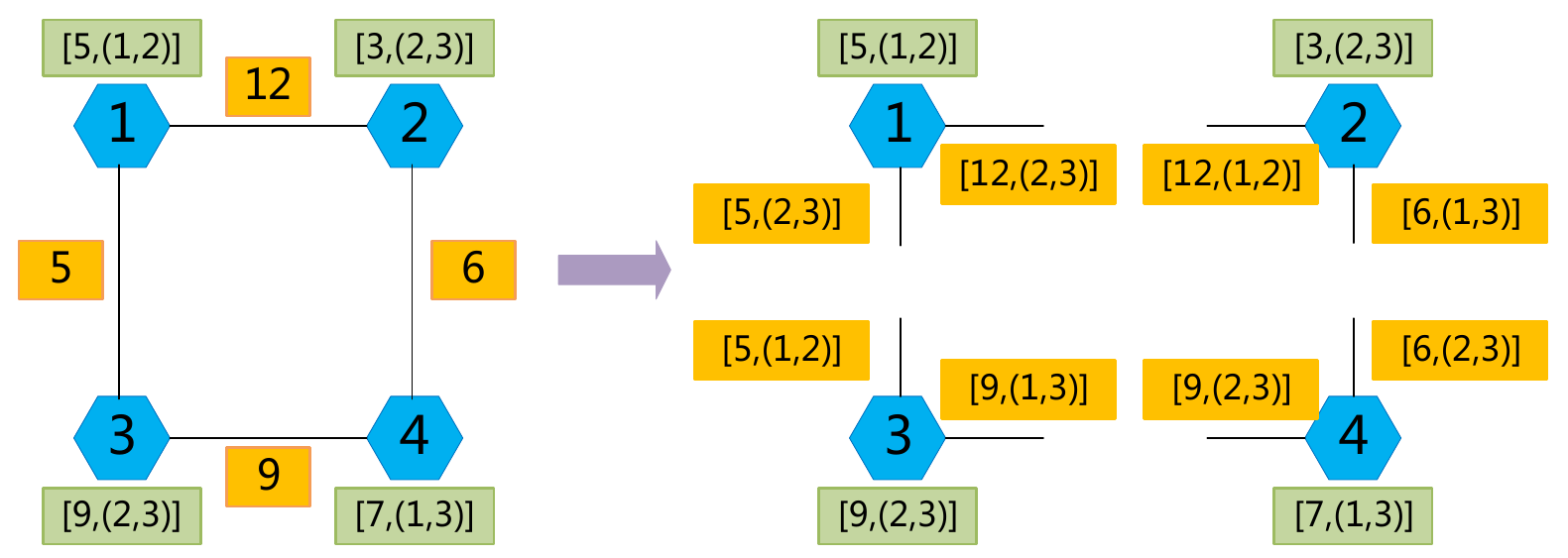}}
\caption{partition cell subgraphs.}
\label{fig partition cell subgraphs}
\end{figure}

The first is the division of the cell subgraph. After receiving the virtual network mapping request from SP, the global controller needs to divide it into cell subgraphs and send it to the local controller. Each virtual node generates a cell subgraph, which includes not only the resource quantity and candidate domain information of the virtual node, but also the candidate domain information of the neighboring nodes and the bandwidth requirements of the links between them. FIGURE \ref{fig partition cell subgraphs} shows the partitioning result of the virtual request subgraph. Each virtual node is individually partitioned. Take node 1 for example. Virtual node 1 has a resource requirement of 5 and can be mapped to candidate domain 1 or candidate domain 2. The two virtual links to node 1 indicate the bandwidth resource requirements. The candidate domain of one link is the candidate domain 2 or 3 of node 2, and the candidate domain of another link is the candidate 2 or 3 of node 4. The same is true for the other three nodes.

After receiving the subgraph information of virtual network request unit, the local controller calculates several candidate nodes of each virtual node in the domain according to the subgraph and in-domain information. Calculate the average mapping cost estimate of physical nodes in each virtual node mapping region according to the following formula, then select the two nodes with the lowest and the next lowest cost as candidate nodes in these schemes.

\begin{equation}
\begin{aligned}
PreCost_{i,j,k}=CPU(n_i^v) \times C(n_k^s)+\frac{\sum_{links}\sum_{candiDomain} BW(l) \times C_{k,b}}{NoL} ,
\end{aligned}
\end{equation}

where $PreCost_{i,j,k}$ represents the estimated average mapping cost of virtual node i to physical node k in physical domain j. The first term $CPU(n_i^v)  \times C(n_k^s)$ represents the product of the CPU requirements of the virtual node i and the Cost of the physical node k.

The second term represents the average virtual link mapping cost in the virtual request subgraph for all possibilities of link candidate domains in the virtual request subgraph. The denominator $NoL$ represents the number of combinations of these link candidate fields. The molecule represents the sum of all possible intra-domain mapping costs of the virtual link in the virtual request subgraph. Because the local controller only knows the link cost within the physical domain. Therefore, the link mapping cost can only be estimated in terms of the cost from that node to the boundary node. The value of $C_{k,b}$ in this formula is divided into three cases for discussion.

\begin{enumerate}
\item If virtual node 1 is connected to virtual node 2 and one of the candidate fields of virtual node 2 is j, then $C_{k,b}$ is 0.
\item If virtual node 1 is connected to virtual node 3, one candidate domain of virtual node 3 is another physical domain m. And this physical domain m is directly connected to the physical domain j. Then, $C_{k,b}$ at this point should be the link cost from node k to the boundary node directly connected to the physical domain m.
\item If virtual node 1 is connected to virtual node 4, a candidate field for virtual node 4 is n. And this candidate domain n is not directly related to this physical domain j. Then $C_{k,b}$ at this point should be the average cost from node k to all boundary nodes.
\end{enumerate}

After the local controller completes the calculation and selection of candidate nodes, the following information is uploaded to the global controller as the basis for integrating pseudo topology and partitioning virtual network mapping requests.

\begin{enumerate}
\item Information of all candidate nodes corresponding to virtual nodes, including their CPU unit price and available CPU resources;
\item The unit price of the lowest cost path between candidate nodes and each boundary node, and the unit price of the lowest cost path between candidate nodes;
\item CPU unit price, time delay, available CPU resources of boundary nodes, and link between boundary nodes and its unit price;
\item Connection of inter-domain links, link unit price, time delay and available resources of links.
\end{enumerate}

After the global controller receives the candidate node scheme and some in-domain information, it integrates a pseudo-topology to represent the cross-domain physical network. Pseudo-topology preserves the topological relations between candidate nodes and boundary nodes in each domain and can be used as particle swarm optimization network topology\cite{Yang2004Overview, Kennedy2010Particle}. PSO algorithm is a kind of stochastic global optimization technique. The algorithm finds the optimal region in the complex search space through the interaction between particles. We introduce genetic variation factors into the traditional PSO algorithm. It can effectively avoid the situation that the particle falls into the local optimum and cannot reach the global optimum. The following is the calculation formula of particle location and velocity of PSO.
\begin{equation}
\begin{aligned}
v_i^{new}=v_i+c_1 rand_1(x_i^{pb}-x_i)+c_2 rand_2(x^{gb}-x_i).
\end{aligned}
\end{equation}
\begin{equation}
\begin{aligned}
x_i^{new}=x_i+v_i^{new}.
\end{aligned}
\end{equation}

In the application scenario of this algorithm, the symbols in the above formula are explained as follows:

\begin{enumerate}
\item $x_i$: The location of the $i$-th particle in PSO refers to the $i$-th mapping scheme in this algorithm. The coordinates of particles in each dimension correspond to which physical node the virtual node is mapped to in the mapping scheme.
\item $v_i$: The velocity of the $i$-th particle in PSO indicates the changing direction of the $i$-th mapping scheme in this algorithm. The speed of the node in each dimension indicates whether and how the virtual node map changes in the node mapping scheme.
\item $x_i^{pb}$: The historical optimal position of the $i$-th particle in particle swarm optimization.
\item $x^{gb}$: Global optimal position in particle swarm optimization.
\item $c_1$, $c_2$: The learning factor in PSO adjusts the maximum stride length of flight to the global best example and individual best example respectively. A reasonable learning factor can accelerate learning without falling into local optimal.
\item $rand_1$, $rand_2$: Random numbers between 0 and 1.
\item Minus sign: The minus sign in the above equation should be defined as 0 if the two operands are the same and 1 if the two operands are different.
\item Plus sign: The plus sign in the above equation should be defined as: add the operands and carry out integer operation. If the addition result is greater than 0.5, then 1 will be obtained. If the addition result is less than 0.5, then 0 will be obtained.
\end{enumerate}

There is the generation of the mapping scheme. According to the final node mapping scheme obtained by PSO, the global controller generates corresponding mapping requests for each physical domain and sends the request to each physical domain. Each request includes node mapping request and link mapping request.

Generate the mapping scheme, the first generation node mapping scheme, which is generated by each virtual node corresponding physical node number. Then according to the node mapping scheme and virtual network mapping request whether there is a virtual link between the virtual nodes is determined whether the link mapping request needs to be generated between the two nodes. The link mapping scheme is generated according to the Floyd shortest path in the pseudo topology.

Floyd algorithm, also known as the interpolation method, is an algorithm that uses the idea of dynamic programming to find the shortest path between multiple source points in the weighted graph. The core idea of this algorithm is to find the shortest path matrix between each two points of a graph by using its weight matrix. Its advantage is that it is easy to understand and calculate the shortest distance between any two nodes\cite{Hao2008Some}.

The local controller of each physical domain maps the nodes and links in the domain according to the received mapping request. Since the global controller only obtains part of the topology information of each physical domain, it is possible that the links in the mapping request are spliced by multiple physical links in the physical domain. Therefore, the issued link mapping request needs to be processed to become a local link mapping scheme. Similarly, the local link mapping scheme is generated from the Floyd shortest path of the local topology.

If there are transit nodes between two physical nodes, the link mapping scheme is generated from the initial node to the destination node in turn. The physical node mapping scheme needs no further processing.

\section{Multi-Domain Virtual Network Mapping Algorithm Implementation}

Candidate node selection algorithm describes the process of candidate node selection by local controller after receiving virtual network mapping request subgraph from global controller. PSO algorithm describes the process that global controller completes the selection of node mapping scheme after receiving the information uploaded by local controller. The mapping scheme generation algorithm describes the process of the global controller generating the corresponding node mapping scheme and link mapping scheme for each physical domain according to the node mapping scheme. The physical network mapping algorithm describes the process that each local controller completes the final node mapping and link mapping according to the issued mapping scheme.

\subsection{Candidate Node Selection}
This module aims to select candidate nodes in its candidate domain for each virtual mapping subgraph. The main principle of the selection of candidate nodes is to select the node with the lowest estimated mapping cost. The overall idea is to calculate and sort the estimated mapping cost of each physical node, then select the physical node which has the lowest estimated mapping cost and has not become a candidate node as the candidate node of the virtual node. Each virtual node selects only one of its candidate nodes in one of its mapped domains.

\begin{algorithm}
  \caption{Candidate Node Selection Algorithm}
  \begin{algorithmic}[1]
     \Require
        {$G_i^s=\{N_i^s,L_i^s\}$, $G_i^v=\{N_i^v, L_i^v\}$};
    \Ensure
        {$\{N_{candi}\}$};
    \For {$sn:substerateNodeList$}
    \State $calculateCost(sn)$;
    \EndFor
    \State $sort(CostOfSubstrateNode)$;
    \While {$findFlag==false$}
    \State $sn=getMinCostNode()$;
    \If {$isMarked(sn)==false$}
    \State $sn.setMarked(true)$;
    \State $setFindFlag(true)$;
    \EndIf
    \EndWhile
    \State \Return $CandiNode$;
  \end{algorithmic}
\end{algorithm}

\subsection{Particle Swarm Optimization}

After collecting all the information uploaded by the physical domain, the global controller integrates the collected topology and resource information into a global pseudo-topology, then carries out particle swarm optimization algorithm on this pseudo-topology. The objective of PSO is to select the mapping scheme with the lowest cost in a limited number of iterations. In this paper, the virtual node mapping scheme is the position of particles in PSO and the particle fitness function is the comprehensive cost of the mapping scheme. Because the general particle swarm optimization algorithm may fall into the local optimum and cannot reach the global optimum, the genetic mutation factor is added to the particle swarm optimization algorithm. The particle has a certain probability to reset its position randomly, which can reduce the probability of falling into local optimum.

\begin{algorithm}
  \caption{Particle Swarm Optimization}
  \begin{algorithmic}[1]
     \Require
        {The $G_i^v=\{N_i^v,L_i^v\}$ and PseudoTopo};
    \Ensure
        {The Embedding Result for $N_v$};
    \State $initialPartical()$;
    \For {$i<iterationTime$}
    \For {$partical:particalList$}
    \State $computeParticalVelocity(partical)$;
    \State $changeParticalPosition(partical)$;
    \State $judgeWhetherToReSet()$;
    \If {$reset()$}
    \State $resetParticalPosition(partical)$;
    \EndIf
    \State $computeParticalCost(partical)$;
    \If {$partical.cost<particalBest.cost$}
    \State $particalBest=partical$;
    \EndIf
    \If {$partical.cost<globalBest.cost$}
    \State $globalBest=partical$;
    \EndIf
    \EndFor
    \EndFor
    \State \Return $globalBest$;
  \end{algorithmic}
\end{algorithm}

\subsection{Mapping Scheme Generation}
The process of generating the mapping scheme is to first determine the virtual node mapping scheme, then generate the virtual link mapping scheme according to the virtual node mapping scheme and the path generated by Floyd algorithm. The mapping scheme is generated for each physical domain, including physical node resource requirements, intra-domain link bandwidth requirements and inter-domain link bandwidth requirements.

\begin{algorithm}
  \caption{Generate Embedding Request}
  \begin{algorithmic}[1]
     \Require
        {The Candidate Node List and PseudoTopo};
    \Ensure
        {The Embedding Request};
    \State $initialRequest()$;
    \For {$cn:CandidateNodeList$}
    \State $addSubstrateNodeToRequest(cn)$;
    \EndFor
    \For {$vrn1:VirtualRequestList$}
    \For {$vrn2:VirtualRequestList$}
    \If {$existVirtualLink(vrn1,vrn2)$}
    \State $addSubstrateLinkToRequest()$;
    \EndIf
    \EndFor
    \EndFor
    \State \Return $EmbeddingRequest$;
  \end{algorithmic}
\end{algorithm}

\subsection{Physical Network Mapping}
The local controller maps the physical nodes and links according to the mapping scheme issued by the global controller, and calculates the mapping cost.  Although the mapping scheme has identified the physical nodes and links to be mapped, the global controller does not have a complete understanding of the physical network topology of each domain, so this mapping scheme needs to be processed by the local controller.

The local controller first determines the physical nodes to be mapped. Then determine whether there are physical links between each pair of physical nodes that need to be mapped. Then the physical nodes and links are mapped according to the requirements of physical links and the paths generated by Floyd algorithm. Finally, the mapping cost is returned.

\begin{algorithm}
  \caption{Physical Network Mapping Algorithm}
  \begin{algorithmic}[1]
     \Require
        {Embedding Request, $G_i^s=\{N_i^s,L_i^s\}$};
    \Ensure
        {The Embedding Cost of Request};
    \State $initialRequest()$;
    \For {$cn:CandidateNodeList$}
    \State $embedSubstrateNode(cn)$;
    \State $computeNodeCost(cn)$;
    \EndFor
    \For {$cn1 : CandidateNodeList$}
    \For {$cn2 : CandidateNodeList$}
    \If {$existRequestLink(cn1,cn2)$}
    \State $embedSubstateLink(cn1,cn2)$;
    \State $computeLinkCost(cn1,cn2)$;
    \EndIf
    \EndFor
    \EndFor
    \State \Return $embeddingCost$;
  \end{algorithmic}
\end{algorithm}

\section{Experimental Results and Analysis}

In order to verify the performance of MP-VNE algorithm proposed in this paper, MP-VNE is compared with MC-VNM algorithm\cite{Geng2016Multi-domain}, VNE-PSO algorithm\cite{Xiao2014Knowledge} and LID-VNE algorithm\cite{Zaheer2010Multi-provider} in various aspects. The description of the four algorithms is shown in TABLE 1. Specifically, these four algorithms will be compared and tested in four aspects: cost, delay, comprehensive cost and reception rate.

\begin{center}
\begin{table}
\centering
\caption{The algorithm description}
\renewcommand\arraystretch{1.5}
\begin{tabular}{|p{15mm}|p{154mm}|}
\hline
Notation & The algorithm description\\
\hline
MP-VNE & Weight is used to convert the multi-objective optimization problem into the single-objective optimization problem. Use the estimated mapping cost formula to select the boundary node. A particle swarm optimization algorithm with genetic variation factors was used to solve the node mapping scheme. \\
\hline
VNE-PSO & The nearest node to the boundary node is selected and the particle swarm optimization algorithm is used to solve the node mapping scheme. \\
\hline
MC-VNM & By using kruskal spanning tree, the link priority map with the lowest link unit price is selected. The link mapping scheme ultimately determines the node mapping scheme.\\
\hline
LID-VNE & Select the mapping target domain for each node based on the quote for each physical domain. Each physical domain is decomposed by the bandwidth demand matrix, and each physical domain maps nodes and links according to the received bandwidth demand matrix. \\
\hline
\end{tabular}
\end{table}
\end{center}

\subsection{Experimental Environment Setting}
The computer used in the simulation is 8GB, 64-bit win10 operating system. The experimental code is written in eclipse using Java. The network topology used in the simulation experiment is generated randomly by eclipse. Parameters of the simulation experiment are shown as follows:

The physical network is divided into four domains. The number of single domain nodes is 30 for each domain, including two boundary nodes. The resources of each physical node are evenly distributed in the range of 100 $\sim$ 300. The cost of physical nodes is evenly distributed from 1 to 10. The physical node delay is evenly distributed on 1 $\sim$ 10. The physical link resources are evenly distributed in the range of 1000 $\sim$ 3000. The physical link cost is evenly distributed on the scale of 1 $\sim$ 10. The physical link delay is evenly distributed between 1 and 10. The cost of inter domain link is uniformly distributed on 5 $\sim$ 15. The delay of inter domain link is uniformly distributed in the range of 10 $\sim$ 30. Physical nodes are connected with each other with a probability of 50\%.

The number of virtual request nodes is 6. The CPU demand of virtual node is evenly distributed on the scale of 1 $\sim$ 10 and the bandwidth demand of virtual link is evenly distributed on the scale of 1 $\sim$ 10. Each virtual network mapping node has two optional mapping domains, and virtual nodes are connected with each other with a probability of 50\%. In the experiment, the arrival time of virtual request simulates the Poisson process. In 100 time units, the number of virtual network requests is subject to the Poisson distribution with an average of 10 and the life cycle of each virtual network is subject to the exponential distribution with an average of 1000 time units.

In particle swarm optimization algorithm, the number of particles is 10 and the number of iterations is 50. Alpha, beta and gamma of particle swarm optimization algorithm are set as 0.3 , 0.3 and 0.4 respectively. In the process of particle swarm optimization, the probability of mutation of particle position is 10\%.

The summary of each parameter is shown in TABLE 2.

\begin{center}
\begin{table}
\centering
\caption{Simulation parameter setting}
\renewcommand\arraystretch{1.5}
\begin{tabular}{|p{40mm}|p{40mm}|}
\hline
Parameter names & Parameter values \\
\hline
physical network domain & 4 \\
\hline
number of nodes & 30 \\
\hline
node resource quantity & U[100, 300] \\
\hline
node cost & U[1, 10] \\
\hline
node delay & U[1, 10] \\
\hline
link resources & U[1000, 3000] \\
\hline
link cost & U[1,10] \\
\hline
link delay & U[1,10] \\
\hline
number of boundary nodes & 2 \\
\hline
link connection rate & 50\% \\
\hline
number of request nodes & 6 \\
\hline
CPU requirements & U[1,10] \\
\hline
bandwidth requirements & U[1,10] \\
\hline
particle number & 10 \\
\hline
number of iterations & 50 \\
\hline
alpha  & 0.3 \\
\hline
beta  & 0.3 \\
\hline
gamma  & 0.4  \\
\hline
mutation probability & 10\% \\
\hline
\end{tabular}
\end{table}
\end{center}

\subsection{Experimental Results and Analysis}

Experiment 1: virtual network request acceptance rate comparison experiment

In order to compare the request acceptance rate of MP-VNE algorithm with that of other algorithms, we conducted a virtual network request acceptance rate comparison experiment.

\begin{figure}[!htp]
\centerline{\includegraphics[width=370pt,height=21pc]{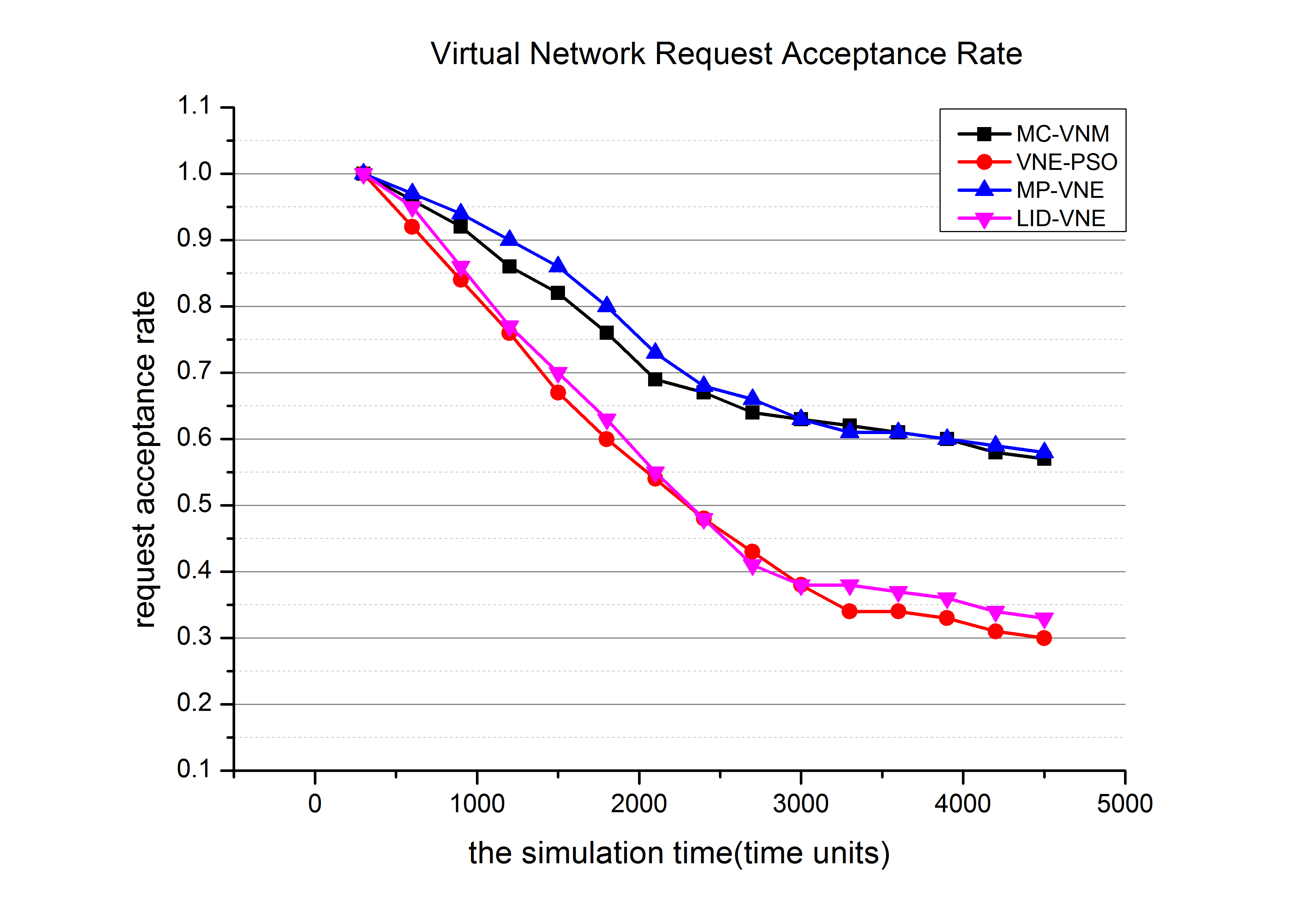}}
\caption{virtual network request reception rate.}
\label{fig_virtual network request reception rate}
\end{figure}

The comparison graph of virtual network request acceptance rate of the four algorithms is shown in the FIGURE \ref{fig_virtual network request reception rate}. With the increase of simulation time, the acceptance rate of virtual network mapping requests of the four algorithms decreases. However, MC-VNM and MP-VNE are better than VNE-PSO and LID-VNE. The MP-VNE algorithms proposed in this paper and slightly better than MC-VNM algorithm, became one of the four algorithms request to accept the highest rate of a virtual network algorithm. It can be seen that MC-VNM and MP-VNE algorithms have a steady trend of receiving rate of virtual network requests around 60\% over time, while the other two algorithms, VNE-PSO and LID-VNE, maintain a downward trend at 30\%.

Experimental results analysis: Virtual network mapping requests require the lowest possible mapping cost, so in the early stage of the cost of nodes and link resources consumption is larger. As a result, quite a few virtual network mapping requests cannot be satisfied, so the early acceptance rate declines rapidly. At a later stage, less costly resources are used up. The remaining virtual network mapping requests can only use expensive resources, where acceptance rates decline slowly.

MP-VNE algorithm has a high receiving rate of virtual network mapping requests, because the algorithm optimizes both cost and delay, and also takes into account the idea of load balancing. As far as possible, the reception rate is as high as possible without losing the first two indicators. However, the other two algorithms perform poorly because the method of selecting mapping nodes is too rigid and rigid, which leads to the shortage of node resources or link resources in the process of receiving virtual network requests. Specifically, VNE-PSO algorithm always chooses the boundary node as the candidate node, so it is easy to cause too much load pressure on the boundary node, resulting in the acceptance rate of virtual network requests is not very ideal. Because the LID-VNE algorithm only knows the average quotation of information in the physical domain, it is easy to map the virtual node to the physical domain with low average quotation, so it will cause excessive load pressure in individual physical domain, resulting in the decline of the receiving rate of virtual network nodes.

Experiment 2: mapping cost comparison experiment

In order to compare the mapping cost between MP-VNE algorithm and other algorithms, we do the mapping cost comparison experiment.

\begin{figure}[!htp]
\centerline{\includegraphics[width=370pt,height=21pc]{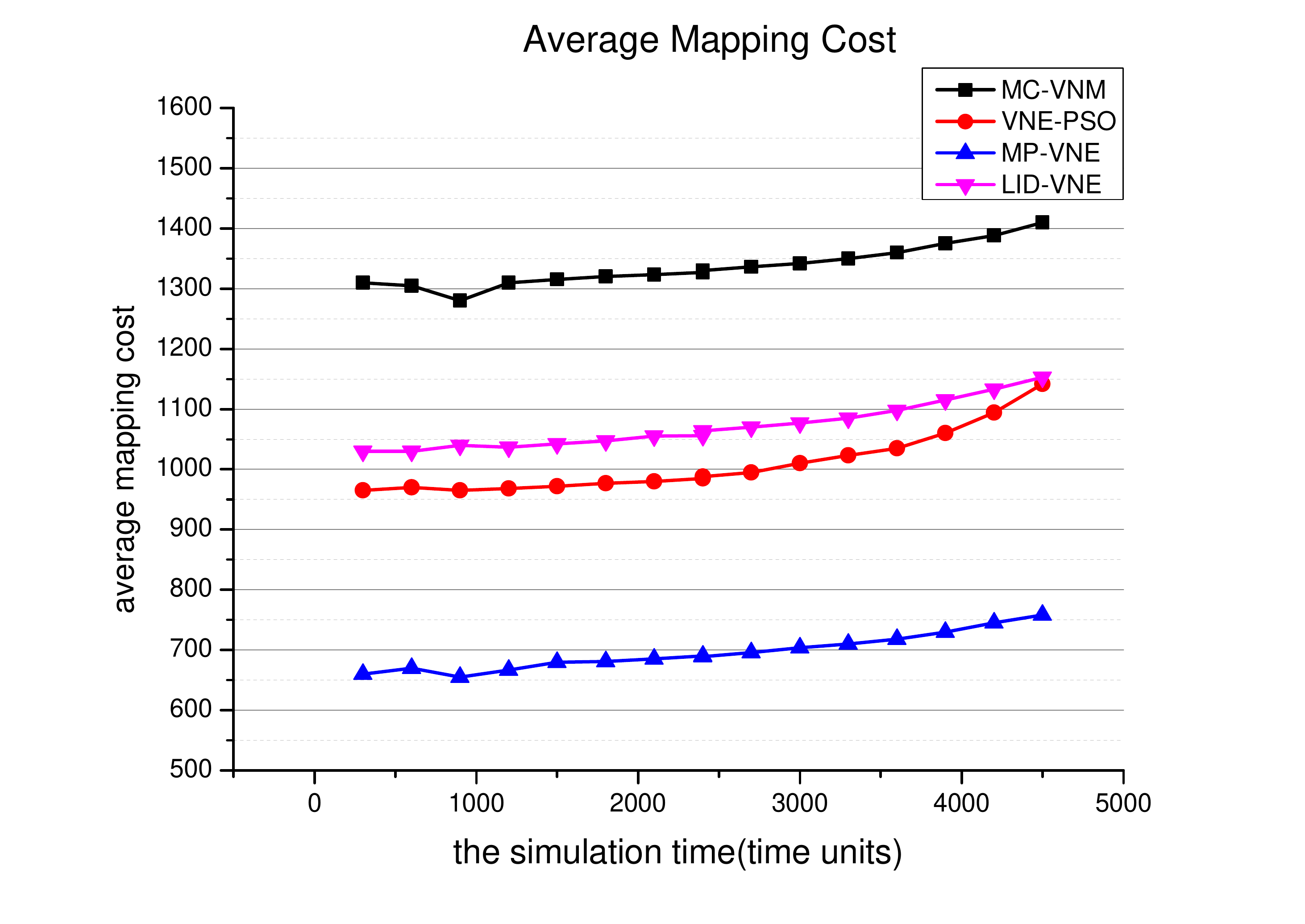}}
\caption{average mapping cost.}
\label{fig_average mapping cost}
\end{figure}

The average mapping cost of the four algorithms is shown in FIGURE \ref{fig_average mapping cost}. Among them, MC-VNM algorithm has the largest average mapping cost, which increases from 1300 to about 1400. The average mapping cost of LID-VNE algorithm is second only to MC-VNM, and increases to about 1150 with time. The average mapping cost of VNE-PSO algorithm ranked third with a large increase. The average mapping cost of MP-VNE algorithm is the smallest, which increases from 650 to about 750.

Experiment results analysis: Relatively, four algorithms of MC-VNM is the worst in terms of the average cost of mapping. This algorithm in solving the problem of NP-hard, did not take the common heuristic algorithm. On the contrary, it uses the traditional optimization method based on Kruskal minimum spanning tree, by the simulation results can be seen that adopted heuristic algorithm optimization effect is obviously better than the traditional mathematical decision method.

LID-VNE algorithm is an algorithm based on distributed virtual network mapping architecture. SP knows little about the network topology and resources of each physical domain, which leads to errors in the decision-making process of virtual network mapping request. After analysis, this algorithm may have a better effect when there is only one candidate physical domain or the physical network topology is small. However, when the network scale is large and the number of virtual network request nodes is large, the shortcomings of this algorithm are exposed. However, this algorithm is still useful when virtual network mapping requests arrive faster, because of the faster decision-making speed. Simulation results show that in pursuit of mapping performance and high requirements for all aspects of mapping results, the centralized virtual network mapping architecture can often achieve better results than the distributed architecture.

VNE-PSO is the most similar algorithm to the MP-VNE algorithm proposed in this paper. Correspondingly, the mapping effect of this algorithm is only second to that of MP-VNE. However, as only the hop number from the boundary node is considered in the selection of candidate nodes, this algorithm does not make a very good decision. The selection method of candidate nodes proposed in this paper is to select candidate nodes by calculating and estimating the mapping cost, which is obviously superior to the simple selection process of candidate nodes in VNE-PSO.

Experiment 3: mapping delay contrast experiment

In order to compare the mapping delay between MP-VNE algorithm and other algorithms, we do the mapping delay comparison experiment.

\begin{figure}[!htp]
\centerline{\includegraphics[width=370pt,height=21pc]{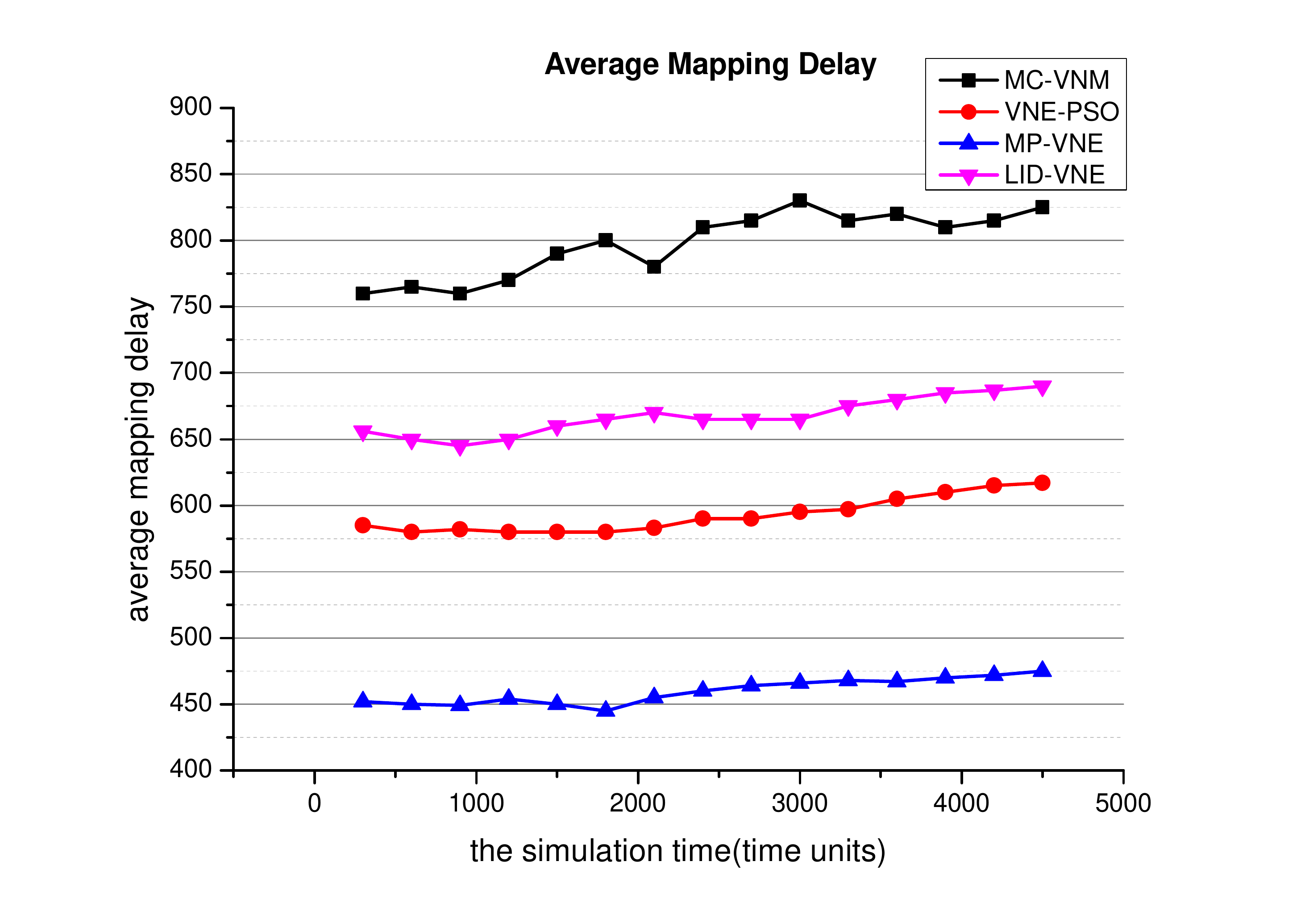}}
\caption{mapping time delay.}
\label{fig_mapping time delay}
\end{figure}

The average mapping delay of the four algorithms is shown in the FIGURE \ref{fig_mapping time delay}. Among them, the average mapping delay of MC-VNM algorithm is the largest, about 800. The LID-VNE algorithm's average mapping delay was second only to MC-VNM, increasing from 650 to about 700. The average mapping delay of VNE-PSO algorithm is the third, about 600. While the average mapping delay of MP-VNE algorithm is the smallest, about 460.

Analysis of experimental results: Among the four algorithms, the average mapping delay of MC-VNM is the largest. Since the other three algorithms are heuristic algorithms, the MC-VNM algorithm uses greedy strategy to map virtual links first and then virtual nodes. It shows that the heuristic algorithm is superior to the simple greedy algorithm in reducing the delay. LID-VNE algorithm is second only to MC-VNM algorithm in delay. This is because the global controller does not know enough about each domain. When the underlying physical network is complex and there are many physical nodes and links, the LID-VNE algorithm cannot select some representative nodes or links for mapping. It can only map all nodes and links as much as possible, so the mapping delay is large. The effect of VNE-PSO algorithm is second only to MP-VNE algorithm. However, because the VNE-PSO algorithm always gives priority to the boundary node as the candidate node, and does not consider the case that the time delay of other nodes is minimal, the process is too rigid. Therefore, it is not as effective as the MP-VNE algorithm proposed in this paper. The MP-VNE algorithm uses the PSO algorithm with genetic variation factors to solve the node mapping scheme, avoiding the situation of falling into the local optimum. Therefore, the best experimental results can be obtained.

Experiment 4: cost of synthetic mapping

In order to compare the comprehensive mapping cost between MP-VNE algorithm and other algorithms, we do a comparative experiment on the comprehensive mapping cost.

\begin{figure}[!htp]
\centerline{\includegraphics[width=370pt,height=21pc]{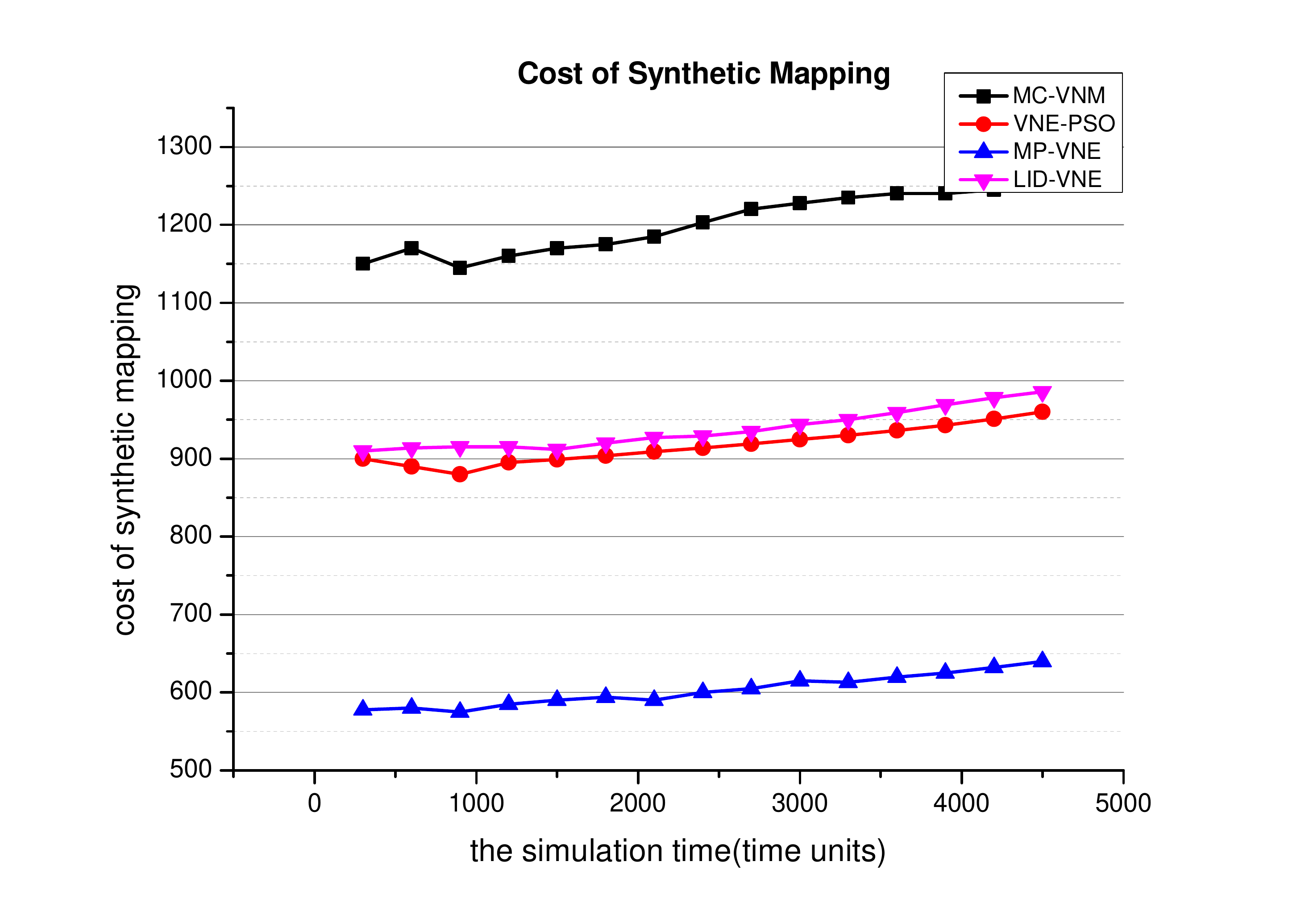}}
\caption{cost of synthetic mapping.}
\label{fig_cost of synthetic mapping}
\end{figure}

In terms of the comprehensive mapping cost, MP-VNE algorithm has the best performance. The highest comprehensive mapping cost is less than 650. The performance of VNE-PSO algorithm is poor, and the comprehensive mapping cost increases to about 950 with time. LID-VNE performs even worse. The cost of synthetic mapping increases to nearly 1000. However, MC-VNM algorithm was the worst performer, with the comprehensive mapping cost increasing from about 1150 to about 1250. These are shown in FIGURE \ref{fig_cost of synthetic mapping}.

Analysis of experimental results: In terms of comprehensive mapping cost, the MP-VNE algorithm is still optimal in terms of comprehensive performance. This is because the algorithm selects candidate nodes by estimating the mapping cost. Each physical domain calculates the estimated mapping cost for all nodes based on the formula that calculates the estimated mapping cost. Then, the node with the least estimated mapping cost is selected as the candidate node, so the overall mapping cost can be reduced. VNE-PSO algorithm is second only to MP-VNE algorithm. However, since only the hop from the boundary node is considered when selecting the candidate node, the mapping cost is not comprehensively considered in this algorithm. LID-VNE algorithm is based on distributed virtual network mapping architecture. The lack of global information leads to errors in deciding which physical domain virtual nodes should be mapped to in the virtual network mapping process. This error leads to an increase in the cost of the mapping. The integrated mapping cost of MC-VNM algorithm is the highest. This is because the algorithm does not adopt the usual heuristic algorithm to solve NP-hard problems. It adopts the traditional optimization method based on Kruskal minimum spanning tree, which results in high cost of synthetic mapping.

\subsection{Summary of This Chapter}
As mentioned above, the MP-VNE algorithm proposed in this paper has good performance in various fields, which is mainly attributed to the following aspects.

First of all, MP-VNE algorithm adopts a centralized hierarchical virtual network mapping architecture, which ensures that there is a global controller under SP to know the general information of the whole network. In the process of selecting candidate mapping domain for each virtual node, we can judge more accurately according to the global information.

Secondly, MP-VNE algorithm adopts the concept of candidate node. Candidate nodes provide additional information in the physical domain, which is also a further guarantee for the global controller to make accurate judgment. In addition, due to the existence of candidate nodes, particle swarm optimization algorithm is easier to converge. When the network scale is large or there are many nodes in the virtual network mapping request, the particle swarm optimization algorithm can obtain better results in a limited number of iterations.

Thirdly, MP-VNE algorithm adopts the particle swarm optimization algorithm with genetic mutation factor. The genetic variation factor ensures that the particles can jump out when they converge to the local optimum, and increases the probability of getting the optimal result. Although the particle swarm optimization algorithm can not guarantee the optimal results in theory, but in the limited computing time, it can often get satisfactory results.

Finally, the link mapping in this paper always follows Floyd algorithm from selection to specific mapping. This algorithm ensures that when the result of node mapping is determined, an optimal path can be selected between physical nodes, so as to minimize the comprehensive index of link delay and bandwidth.

\section{Conclusion}

\subsection{Main Work}
Based on the analysis of previous work, this paper proposes a multi domain virtual network mapping algorithm MP-VNE, which is based on multi-objective optimization. Finally, a simulation experiment is carried out, and the performance of the algorithm proposed is compared with those in other literatures \cite{Geng2016Multi-domain,Xiao2014Knowledge,Zaheer2010Multi-provider}. The performance gap is explained and several key characteristics of the algorithm are obtained. This will guide the design of multi domain virtual network mapping algorithm.

\subsection{Summary of Main Innovation Points}
MP-VNE algorithm proposed in this paper combines multi-objective optimization and multi-domain virtual network mapping algorithm for the first time. In the process of multi-domain virtual network mapping, many factors such as cost and delay are considered simultaneously. In addition, the algorithm adopted the centralized hierarchical multi-domain mapping virtual network architecture, each physical domain of the local controller to select the candidate nodes and uploaded to the global controller, which makes the physical node map. The decision-making process of the global controller for each domain has certain understanding of the basis of the physical network topology, so you can get more optimal mapping results. In addition, the genetic variation factor is added into the PSO algorithm, which can effectively avoid the situation that the particle falls into the local optimal and cannot reach the global optimal in the optimization process.

\subsection{Future Research Direction}
As mentioned before, there are two mainstream solutions to the multi-objective optimization problem. One is a method that converts multi-objective into single objective, the other is Pareto method. The former is adopted in this paper, but the optimization effect of the latter is unknown. In future, we intend to use Pareto method to solve the multi-objective optimization problem and then compare the effect of the two methods.

Many kinds of virtual network mapping models are proposed by heuristic algorithms\cite{Cai2010Virtual, Gu2016Cross, Jiang2018Cross-domain}. One of the most common is greedy algorithm. However, due to the complexity and diversity of the underlying network topology, the greedy strategy cannot obtain the optimal results in many cases. Improving virtual network mapping model or exploring heuristic algorithm to solve virtual network mapping problem is a feasible research direction in the future.

At present, with the rise of network analysis, the application of machine learning and artificial intelligence algorithms in the field of virtual network mapping has become a very popular research direction. For example, genetic algorithms have been applied to multi-domain virtual network mapping algorithms. Machine learning algorithm training model can often make the algorithm achieve relatively good mapping results in a short time, thereby greatly improving the mapping efficiency. In addition, the IoD architecture will be further studied to further improve the scalability and autonomy of the IoD architecture.

\section*{Acknowledgements}
The authors gratefully acknowledge the helpful comments and suggestions of the reviewers, which have improved the presentation.

\nocite{*}
\bibliography{reference}

\section*{Biographies}

\begin{biography}{\includegraphics[width=1in,height=1.25in,clip,keepaspectratio]{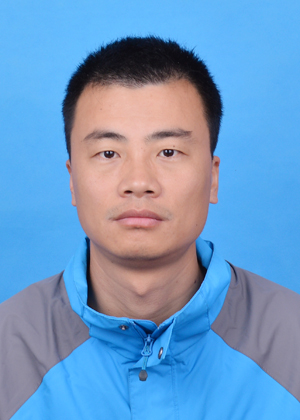}}{\textbf{Peiying Zhang}
is currently an Associate Professor with the College of Computer Science and Technology, China University of Petroleum (East China). He received his Ph.D. in the School of Information and Communication Engineering at the Beijing University of Posts and Telecommunications in 2019. His research interests include semantic computing, future internet architecture, network virtualization, and artificial intelligence for networking. He has published more than 50 papers in prestigious peer-reviewed journals and conferences.}
\end{biography}
~\\
\begin{biography}{\includegraphics[width=1in,height=1.25in,clip,keepaspectratio]{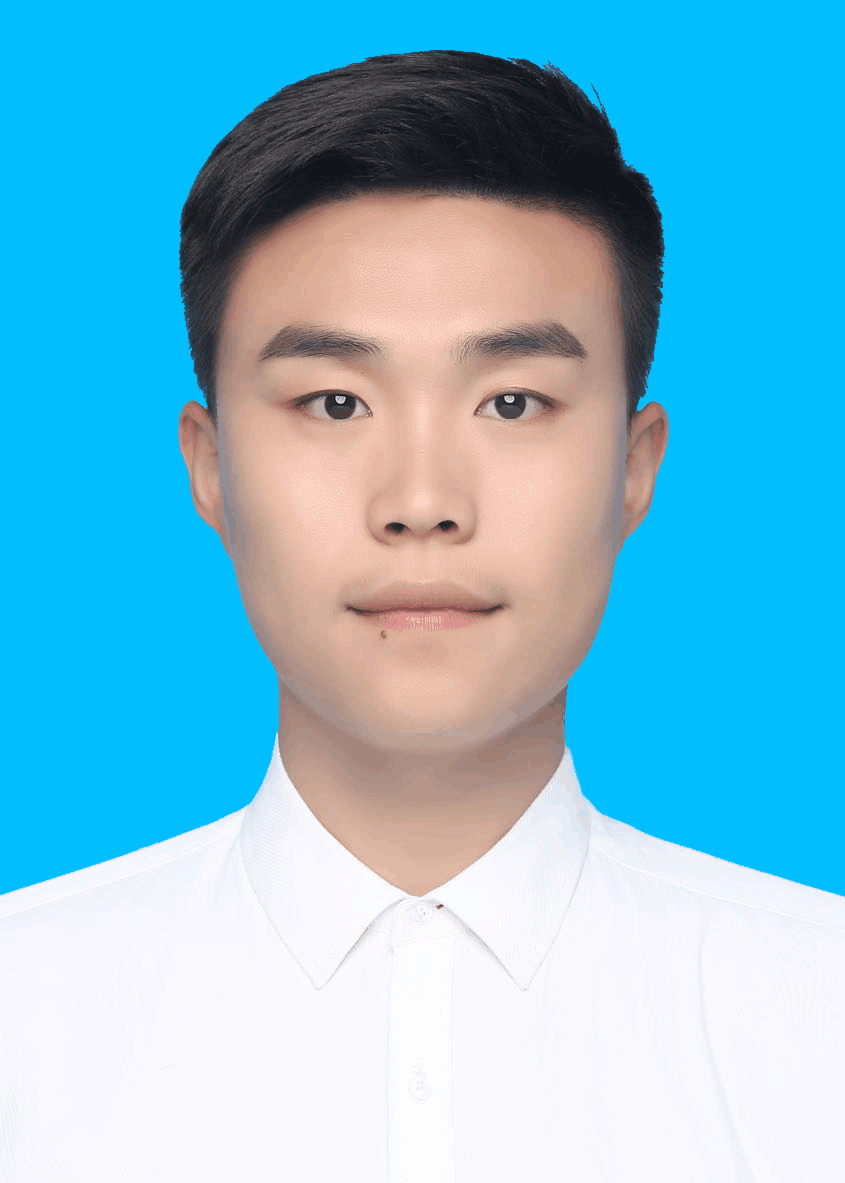}}{\textbf{Chao Wang}
is a graduate student in the College of Computer Science and Technology, China University of Petroleum (East China). His research interests include network artificial intelligence and network virtualization.}
\end{biography}
~\\
~\\
~\\
~\\
\begin{biography}{\includegraphics[width=1in,height=1.25in,clip,keepaspectratio]{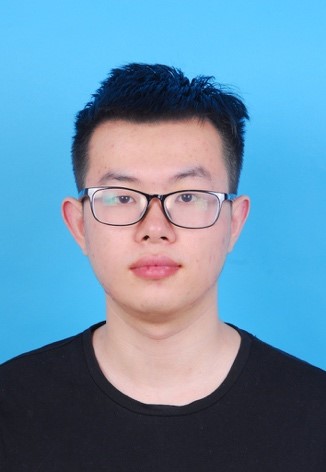}}{\textbf{Zeyu Qin}
is a graduate student in the School of Information and Communication Engineering from the Beijing University of Posts and Telecommunications. His research interests include future network architecture, machine learning and the network virtualization.}
\end{biography}
~\\
~\\
~\\
~\\
~\\
~\\
~\\
~\\
\begin{biography}{\includegraphics[width=1in,height=1.25in,clip,keepaspectratio]{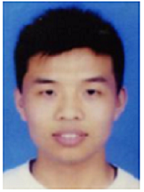}}{\textbf{Haotong Cao}
(S'17) received B.S. Degree in Communication Engineering from Nanjing University of Posts and Telecommunications (NJUPT) in 2015. He is currently pursuing his Ph.D. Degree in NJUPT, Nanjing, China. He was a visiting scholar of Loughborough University, U.K. in 2017. He has served as the TPC member of multiple IEEE conferences, such as IEEE INFOCOM, IEEE ICC, IEEE Globecom. He is also serving as the reviewer of multiple academic journals, such as IEEE Internet of Things Journal, IEEE/ACM Transactions on Networking, IEEE Transactions on Network and Service Management and (Elsevier) Computer Networks. He has published multiple IEEE Trans./Journal/Magazine papers since 2016. His research interests include wireless communication theory, resource allocation in wired and wireless networks. He has received the 2018 Postgraduate National Scholarship of China. He has been awarded 2019 IEEE ICC SecSDN Workshop Best Paper Award.}
\end{biography}

\end{document}